\documentclass{bmvc2k}

\usepackage{enumerate}
\usepackage{algorithm}
\usepackage{algpseudocode}
\usepackage{multirow}
\usepackage{booktabs}
\usepackage{tabularx}
\usepackage{algorithm} 
\usepackage{algpseudocode} 
\usepackage{amssymb}
\usepackage{sidecap}
\usepackage{lipsum}

\title{Structured Probabilistic Pruning for Convolutional Neural Network Acceleration}

\addauthor{Huan Wang*\footnotemark[1]}{huanw@zju.edu.cn}{1}
\addauthor{Qiming Zhang*}{qmzhang@zju.edu.cn}{1}
\addauthor{Yuehai Wang$\dagger$}{wyuehai@zju.edu.cn}{1}
\addauthor{Haoji Hu$\dagger$}{haoji_hu@zju.edu.cn}{1}

\addinstitution{
 College of Information Science and Electronic Engineering\\
 Zhejiang University\\
 Hangzhou, China
}

\runninghead{Huan, Qiming, Yuehai, Roland}{SPP for CNN Acceleration}

\def\eg{\emph{e.g}\bmvaOneDot}

\footnotetext[1]{\noindent * Euqal contribution. $\dagger$ Corresponding author.}
\begin{document}

\maketitle

\begin{abstract}
In this paper, we propose a novel progressive parameter pruning method for Convolutional Neural Network acceleration, named \emph{Structured Probabilistic Pruning (SPP)}, which effectively prunes weights of convolutional layers in a probabilistic manner. Unlike existing deterministic pruning approaches, where unimportant weights are permanently eliminated, SPP introduces a pruning probability for each weight, and pruning is guided by sampling from the pruning probabilities. A mechanism is designed to increase and decrease pruning probabilities based on importance criteria in the training process. Experiments show that, with $4\times$ speedup, SPP can accelerate AlexNet with only $0.3\%$ loss of top-5 accuracy and VGG-16 with $0.8\%$ loss of top-5 accuracy in ImageNet classification. Moreover, SPP can be directly applied to accelerate multi-branch CNN networks, such as ResNet, without specific adaptations. Our $2\times$ speedup ResNet-50 only suffers $0.8\%$ loss of top-5 accuracy on ImageNet. We further show the effectiveness of SPP on transfer learning tasks.
\end{abstract}

\section{Introduction}
\label{sec:introduction}
Convolutional Neural Network~(CNN) has obtained better performance in classification, detection and segmentation tasks than traditional methods in computer vision. However, CNN leads to massive computation and storage consumption, thus hindering its deployment on mobile and embedded devices.


\begin{figure}
   \begin{minipage}{0.5\linewidth}
      \centering
      \includegraphics[width=1.0\textwidth]{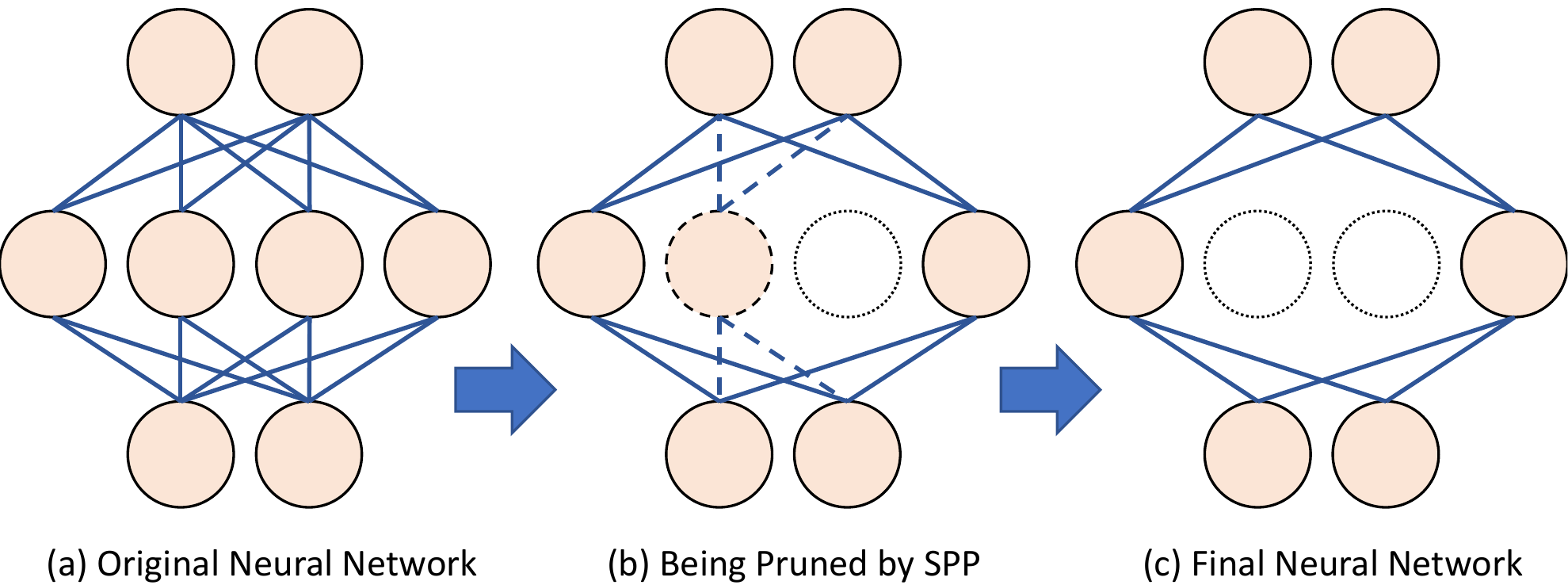}
      \caption{\textbf{left}: The main idea of probabilistic pruning. We assign different pruning probabilities $p$ to different neurons/weights based on some criterion. The dashed circle/line means the neuron/weight has not been totally pruned ($0<p<1$); while the blank circle means the neuron has been eliminated ($p=0$), and thus corresponding connections removed. \textbf{right}: Three kinds of sparsity structure. In im2col implementation, convolutional kernels are expanded into weight matrices. Dark squares mean the pruned weights.}
   \end{minipage}
   \begin{minipage}{0.5\linewidth}
      \centering
      \includegraphics[width=1.0\textwidth]{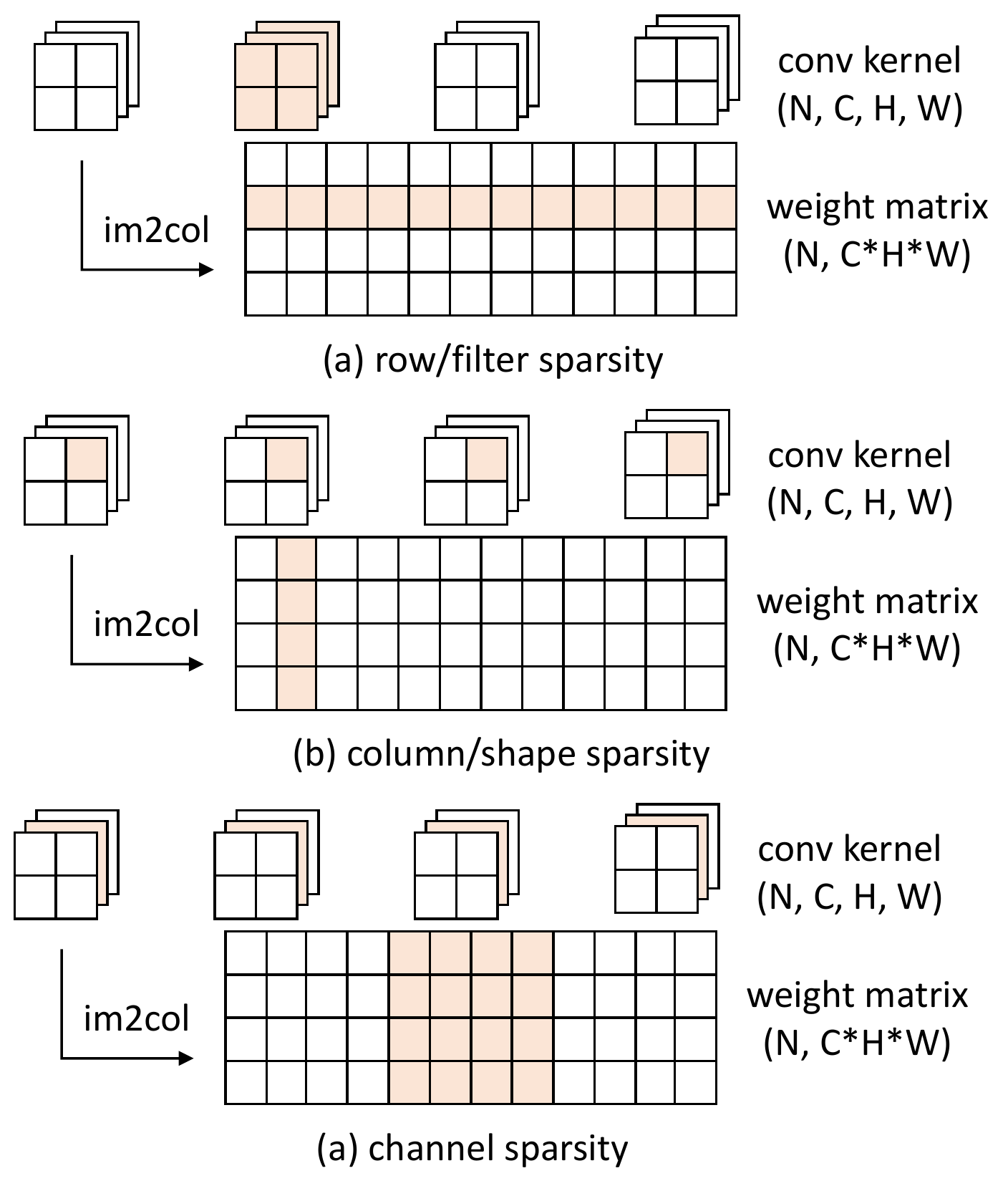} 
   \end{minipage}
   \label{fig:SPP_sparsity-structure}
\end{figure}

Pruning is a promising way for CNN acceleration which aims at eliminating model parameters based on a performance loss function. However, unstructured pruning will lead to irregular sparsity, which is hard to implement for speedup on general hardware platforms~\cite{HanLiuMao16}. Even with sparse matrix kernels, the speedup is very limited~\cite{WenWuWan16}. To solve this problem, many works focus on structured pruning, which can shrink a network into a thinner one so that the implementation of the pruned network is efficient~\cite{AnwSun16,SzeCheYanEme17}. For example, many works were proposed to prune the filters, columns, or the channels of convolutional kernels~\cite{LiKadDurEtAl17,MolTyrKar17,yu2017accelerating,WenWuWan16,guo2016dynamic}, illustrated in Fig.\ref{fig:SPP_sparsity-structure} (right).


However, existing importance-based pruning approaches mainly have a problem: they prune unimportant weights based on some importance criteria and never recover them in the following training process. Given the importance criteria are either simple, such as the commonly used $L_1$ norm and $L_2$ norm~\cite{HanTra15,LiKadDurEtAl17}, or derived under strong assumptions, such as the parameters need to be independent and identically distributed~\cite{MolTyrKar17}, it is likely that some pruned weights may become important later if they were kept through the whole training process.  Therefore, it is necessary to design recovery mechanisms for pruned weights to correct the misjudgments during early training stages.

To solve this problem, we propose the \emph{Structured Probabilistic Pruning~(SPP)} for CNN acceleration, which prunes weights in a \emph{probabilistic} manner, as shown in~Fig.\ref{fig:SPP_sparsity-structure} (left). We assign a pruning probability to each weight. When some weights are below the importance threshold and should have been pruned, we only increase its pruning probability rather than totally eliminate them. Only when $p$ reaches $1$ will the weights be permanently eliminated from the network. We also design a mechanism to decrease the pruning probability if the weights become more important during training, thus correcting the previous misjudgments. Moreover, SPP prunes the whole network at the same time instead of layer-wisely, so the time complexity is controllable when network becomes deeper.



\textbf{Related Work.} 
Intensive research has been carried out in CNN acceleration, which is normally categorized into the following five groups. 
(1) The direct way is to design a more compact network. For example, SqueezeNet~\cite{IanMosAsh16}, MobileNet~\cite{howard2017mobilenets}, ShuffleNet~\cite{zhang2017shufflenet} were proposed to target the mobile devices. 
(2) Parameter quantization reduces CNN storage by vector quantization in the parameter space. \cite{HanMaoDal15} and \cite{WuLenWanHuChe16} used vector quantization over parameters to reduce redundancy. As the extreme form of quantization, binarized networks were proposed to learn binary value of weights or activation functions~\cite{CouBen16,LinCouMemBen16,RasOrdRedFar16,hu2018hashing}. 
(3) Matrix decomposition modifies weights into smaller components to reduce computation. For example, several methods based on low-rank decomposition of convolutional kernel tensor were also proposed to accelerate the convolutional layer~\cite{DenZarBruLecFer14,JadVedZis14,LebYarRakOseLem16,ZhaZouHeSun16}. 
(4) Parameter pruning was pioneered in the early development of neural networks. Optimal Brain Damage~\cite{CunDenSol90} leveraged a second-order Taylor expansion to select parameters for deletion. Deep Compression~\cite{HanMaoDal15} removed close-to-zero connections and quantized the remained weights for further compression. Structured Sparsity Learning~\cite{WenWuWan16} used group LASSO regularization to prune weight rows or columns. Taylor Pruning~\cite{MolTyrKar17} used a Taylor expansion based importance criteria to prune filters. Filter Pruning~\cite{LiKadDurEtAl17} was an one-shot pruning method, using the~$L_1$~norm to prune filters. Recently, Channel Pruning~\cite{He2017Channel} alternatively used LASSO regression based channel selection and feature map reconstruction to prune filters. Further,~\cite{he2018adc} employed Reinforcement Learning to automatically learn the pruning ratio for different layers. Besides, recent works~\cite{molchanov2017variational,neklyudov2017structured,louizos2017bayesian} applied dropout to neural network sparsification by interpreting it in a Bayesian manner. 
(5) Some other works seek acceleration from the perspective of arithmetic complexity, since the acceleration of CNNs are strongly dependent on the implementation of convolution operation. The mainstream implementation is \verb+im2col+, which transforms the convolution into matrix multiplication~(Fig.\ref{fig:alexnet-sensitivity_conv1-visualization} right)~\cite{ChePurSim06,CheWooVan14}. There are also other implementations like Winograd~\cite{DBLP:journals/corr/Lavin15b} and FFT~\cite{DBLP:journals/corr/VasilacheJMCPL14,vasilache2014fast}, which are faster theoretically but can cause side-effects such as bandwidth bottlenecks in practice.

\section{The Proposed Method}
\label{sec:proposed}
Suppose that the we have a dataset~$D$ consisting of~$N$ inputs and their corresponding labels: $D=\left\{(\boldsymbol{X}_1,y_1), (\boldsymbol{X}_2,y_2),\dots, (\boldsymbol{X}_N,y_N)\right\} \nonumber.$ The parameters of a CNN with~$K$ convolutional layers is represented by $\Omega = \big\{(\boldsymbol{W}^1, \boldsymbol{b}^1), (\boldsymbol{W}^2, \boldsymbol{b}^2), ..., (\boldsymbol{W}^K, \boldsymbol{b}^K) \big\}, \nonumber $ which are learned to minimize the discrepancy, i.e.~the loss function~$L(D|\Omega)$, between network outputs and labels. The common loss function for classification tasks is the negative log-likelihood of Softmax output~$q$, which is defined as $L(D|\Omega) = -\frac{1}{N}\sum_{j=1}^N \log q_{y_j}$, where~$q_{y_j}$ represents the~$y_j$th element of the Softmax output for the $j$th input.



The aim of parameter pruning is to find a simpler network~$\Omega'$ with fewer convolutional parameters based on the original network~$\Omega$, in which the loss is minimized. This minimization problem is defined by Eqn.(\ref{eqn:optimization_formula}).
\begin{equation}
   \min \limits_{\Omega'}  \quad L(D|\Omega') \quad \text{s.t.} \quad ||\Omega'||_0 < ||\Omega||_0
\label{eqn:optimization_formula}
\end{equation}

Normally for CNN, an input tensor $\boldsymbol{Z}^l \in \mathbb{R}^{c^l \cdot h^l \cdot w^l}$ of convolutional layer $l \in \{1, 2, ..., K\}$ is firstly convolved with the weight tensor~$\boldsymbol{W}^l \in \mathbb{R}^{c^{l+1} \cdot c^l \cdot \hat{h}^l \cdot \hat{w}^l}$, then a non-linear activation function $f(\cdot)$, usually Rectified Linear Units~(ReLU), will be applied to it. Then the output will be passed as input to the next layer. For pruning, a mask~$g\in \{0,1\}$ is introduced for every weight, which indicates whether this weight is used in the network. Thus, the output of the $l$th layer is described as
\begin{equation}
   f((\boldsymbol{g}^l \odot \boldsymbol{W}^l) * \boldsymbol{Z}^l + \boldsymbol{b}^l),
\label{eqn:mask_formula}
\end{equation}
where $(\odot)$ denotes element-wise multiplication and~$(*)$ denotes the convolution operation. Note that the masked weights are also not updated during back propagations. In the context of structured pruning, weights are divided into different groups. Technically, a weight group can be defined in any way, but in practice weights are usually grouped in a way friendly to hardware implementation, \eg~the weights in the same row, column or channel are called a weight group (see Fig.\ref{fig:SPP_sparsity-structure} right). Since SPP targets structured pruning of convolutional layers, we assign the same~$g$ to all weights in the same weight group, which are pruned or retained simultaneously in each iteration.

For traditional pruning methods, once weights are pruned, they will never be reused in the network, which ignores the interdependency and plasticity of the neural network, because during pruning some initially unimportant weights may become important later due to the missing of other dependent weights. We term this kind pf pruning as~\emph{deterministic pruning}. To address this problem, some buffering mechanism to postpone pruning is in demand so that weights can be evaluated more times before some of them are decidedly deemed as unimportant and get removed permanently. Inspired by dropout~\cite{HinSriKri12}, we reinterpret pruning in a probabilistic manner, termed as \emph{probabilistic pruning}. Specifically, a pruning probability~$p$ is introduced for each weight. For example, $p = 0.7$ means that there is $70\%$ likelihood that the mask~$g$ of the corresponding weight is set to zero.  During the training process, we increase or decrease all~$p$'s based on the importance criterion of weights. Only when~$p$ is increased to~$1$, its corresponding weight is permanently removed from the network. Obviously, deterministic pruning can be regarded a specific case of probabilistic pruning when we choose to increase $p$ from~$0$ to~$1$ in a single iteration.

\textbf{How to update~$p$.}
Assume that a convolutional layer consists of~$N_c$ weight groups. Our aim is to prune~$R N_c$ among them, where $R \in (0,1)$ is the pruning ratio, indicating the fraction of weight groups to be pruned at the final stage. SPP updates~$p$ by a competition mechanism based on some  importance criterion. In this paper, we choose the importance criterion as the~$L_1$ norm of each weight group: the bigger the~$L_1$ norm, the more important that weight group is. In our experiments, we find that~$L_1$ and~$L_2$ norms have similar performance as pruning criterion, in line with previous works~\cite{HanTra15,LiKadDurEtAl17}. There are also other importance criteria such as Taylor expansion~\cite{CunDenSol90,MolTyrKar17} to guide pruning. In this paper, we choose the~$L_1$ norm for simplicity, but our method can be easily generalized to other criteria.

The increment~$\Delta$ of~$p$ is a function of the rank~$r \in \{0, 1, \dots, N_c-1\}$, where the rank~$r$ is obtained by sorting the~$L_1$ norm of weight groups in ascending order. It may seem natural to use $L_1$ norm rather than the rank of $L_1$ norm as criterion, but in practice we find rank is more controllable (it plays a role like smoothing) and gives better results, so we tend to use the rank as criterion. The function~$\Delta(r)$ should satisfy the following two properties: (1) $\Delta(r)$ is a strictly decreasing function because higher rank means greater~$L_1$ norm. In this situation, the increment of pruning probability should be smaller since weights are more important based on the~$L_1$ norm importance criterion. (2) $\Delta(r)$ should be zero when~$r=R N_c$. Since we aim at pruning~$R N_c$ weight groups at the final stage, we need to increase the pruning probability of weights whose ranks are below~$R N_c$, and decrease the pruning probability of weights whose ranks are above~$R N_c$. By doing this, we can ensure that exactly~$R N_c$ weight groups are pruned at the final stage of the algorithm.

\begin{figure}
   \begin{minipage}[t]{0.5\linewidth}
      \centering
      \includegraphics[width=0.81\textwidth]{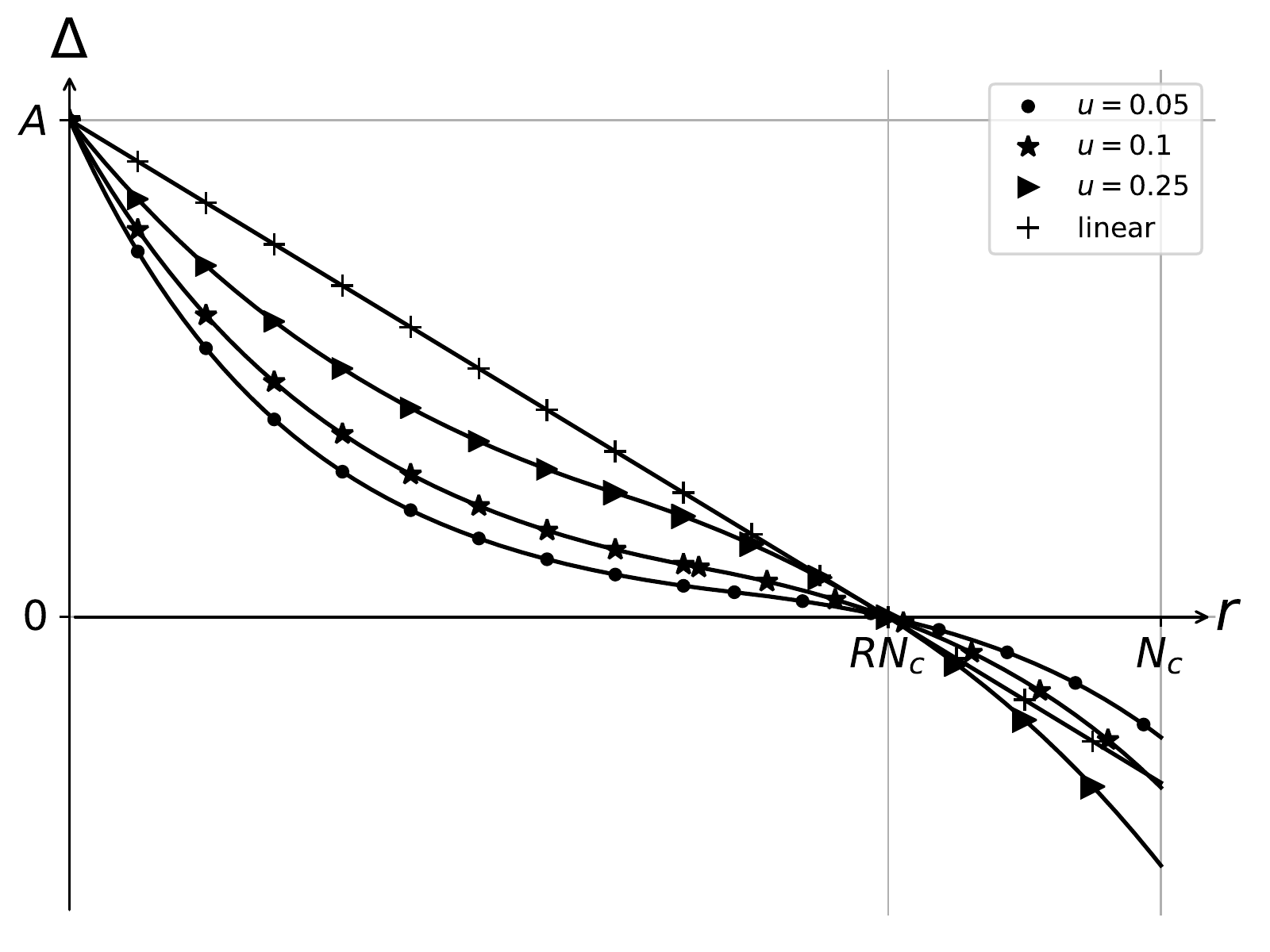}
   \end{minipage}
   \begin{minipage}[t]{0.5\linewidth}
      \centering
      \includegraphics[width=1.0\textwidth]{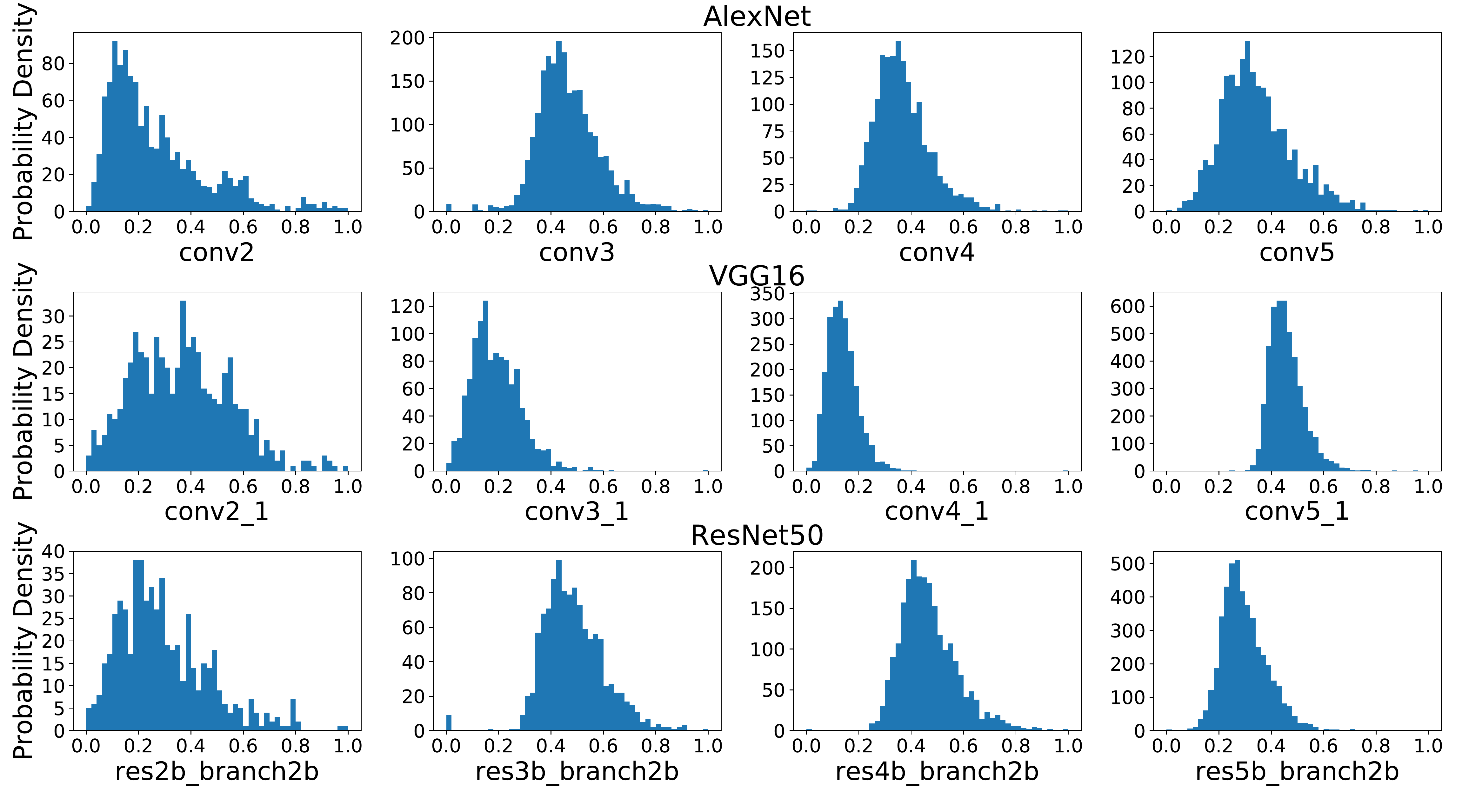} 
   \end{minipage}
   	\caption{\textbf{left}: Functional relationship of the pruning probability increment~$\Delta$ with regard to the rank~$r$. Straight line is the linear function, the others comes from proposed center-symmetric exponential function. \textbf{right}: The~$L_1$ norm histogram of convolutional layers in AlexNet, VGG-16 and ResNet-50. Four representative layers are shown here.}
	\label{fig:punish_density}
\end{figure}

The simplest form satisfying these properties is a linear function, as shown in Fig.\ref{fig:punish_density} (left). However, our experiments show that it is not very good if we take~$\Delta(r)$ as a linear function. The reason, we think, is that the~$L_1$ norm is used to measure the weight importance, but the~$L_1$ norm is not uniformly distributed. In Fig.\ref{fig:punish_density} (right), we plot the~$L_1$ norm histogram of convolutional layers in AlexNet, VGG-16 and ResNet-50. It is observed that the~$L_1$ norm of each layer shares a Gaussian-like distribution in which the vast majority of~$L_1$ values are accumulated within a very small range. If we set~$\Delta(r)$ linearly, the variations of increments would be huge for middle-ranked weight groups, but their actual~$L_1$ values are very close. Intuitively, we need to set~$\Delta(r)$ of middle-ranked columns similar and make~$\Delta(r)$ steeper on both ends, making it in correspondence with the distribution of~$L_1$~norms. So we propose a center-symmetric exponential function to achieve this goal, as shown in Eqn.(\ref{eqn:Delta_i}).

\begin{equation}
\Delta(r) = \left\{
      \begin{aligned}
         &Ae^{-\alpha r}        &,& \text{\emph{ if} } r \leq N \\
         &2uA - Ae^{-\alpha(2N-r)}          &,& \text{\emph{ if} } r>N
      \end{aligned}
    \right.
\label{eqn:Delta_i}
\end{equation}
Here~$u \in (0,1)$ is a hyper-parameter to control the flatness of the function, smaller~$u$ indicates that~$\Delta(r)$ for middle-ranked weight group is flatter, as shown in Fig.\ref{fig:punish_density} (left). Note that~$\alpha$ is a decaying parameter for the exponential function, and the function is center-symmetric on point~$(N,uA)$. Since we need to compel~$\Delta(r)$ passing through~$(N,uA)$ and~$(RN_c,0)$, we can actually solve out~$\alpha$ and~$N$ as~$\alpha=\frac{\log(2)-\log(u)}{RN_c}$ and~$N=-\frac{\log(u)}{\alpha}$.

For each weight group, its pruning probability is updated by
\begin{equation}
   p_{k+1} = \max(\min(p_k+\Delta(r_k),1),0)
   \label{eqn:update_p}
\end{equation}
where~$k$ denotes the~$k$th update of~$p$, the min/max is to ensure that~$p$ is within the range of~$[0,1]$.  During the~$k$th update, for a specific weight group, if its rank~$r_k$ is less than~$RN_c$, $\Delta(r_k)$ would be positive to make~$p_k$ increase; and if its rank~$r_k$ is greater than~$RN_c$, $\Delta(r_k)$ would be negative to make~$p_k$ decrease.

\textbf{When to update $p$.}
A common practice is to prune weights at a fixed interval~\cite{MolTyrKar17}. In SPP, we follow this simple rule: $p$ is updated every~$t$ iterations of training, where~$t$ is a fixed integer. For SPP training, after all the $p$'s are updated, the masks~$g$'s are generated by Monte Carlo sampling according to~$P(g=0) = p.$ After~$g$ is obtained for each weight, the pruning process is applied by~Eqn.(\ref{eqn:mask_formula}). Finally, after the pruning ratio~$R$ is reached, we stop pruning and retrain the pruned model for several epochs to compensate for accuracy. The whole algorithm of SPP is summarized in Algorithm~\ref{alg:SPP}. Note that convolutional layers own little percentage of parameters in the whole network~\cite{KriSutHin12,Simonyan2014Very}, so masks assigned to weights in convolutional layers will cause little burden to the memory. Also they are very easy to implement, since there are off-the-shelf optimized GPU multiplication kernels to help apply masks to weights.


\begin{algorithm}[ht]
\caption{The SPP Algorithm}
\begin{algorithmic}[1]
\State Input the training set~$D$, the original pre-trained CNN model~$\Omega$ and target pruning ratio~$R$.
\State Set hyper-parameters $A$, $u$ and $t$ (in default, $A=0.05$, $u=0.25$ and~$t=180$).
\State Set update number~$k=0$.
\State For each weight group in all conv layers, initialize its pruning probability~$p_k=0$.
\State Initialize iteration number~$i=0$.
\Repeat
    \State If $\mod(i, t)=0$, then update $p_k$ by Eqn.(\ref{eqn:Delta_i}) and~(\ref{eqn:update_p}), and set~$k=k+1$.
    \State For each weight group, obtain~$g_i$ by Monte Carlo sampling based on~$p_k$.
    \State Prune the network by Eqn.(\ref{eqn:mask_formula}).
    \State Train the pruned network, updating weights by back propagations.
    \State $i=i+1$.
\Until {The weight group ratio of~$p_k=1$ equals to~$R$.}
\State Retrain the pruned CNN for several iterations.
\State Output the pruned CNN model~$\Omega'$.
\end{algorithmic}
\label{alg:SPP}
\end{algorithm}

\section{Experiments}
The hyper-parameters in SPP are~$A$, $u$ and~$t$. Here we set~$A=0.05$, $u=0.25$ and~$t=180$. In practice, we found the proposed algorithm is quite robust to these hyper-parameters across different neural network architectures and datasets, so actually the \emph{same} hyper-parameter setting is used for \emph{all} experiments. Other settings such as weight decay, momentum, and dropout are unchanged from the baseline model. In the pruning and retraining process, we only adjust the learning rate and batch size, which will be elaborated in the following experiments. As for sparsity structure, considering a filter typically contains much more weights than a column does, in this paper we choose to prune \emph{columns} of the weight matrices (see Fig.\ref{fig:SPP_sparsity-structure} right), which we believe will cause less severe damage to network in pruning.

On the small-scale dataset CIFAR-10~\cite{KriHin09}, we evaluate our method on a shallow model ConvNet~\cite{KriSutHin12}. Then on the large-scale dataset ImageNet-2012~\cite{DenDonSocEtAl09}, we evaluate our method with three popular models, i.e.~AlexNet~\cite{KriSutHin12}, VGG-16~\cite{Simonyan2014Very} and ResNet-50~\cite{HeZhaRenSun16}. We use \verb+Caffe+~\cite{JiaSheDonEtAl14} for all of our experiments. Methods for comparison include four structured pruning approaches which were proposed in recent years: Structured Sparsity Learning (SSL)~\cite{WenWuWan16}, Taylor Pruning (TP)~\cite{MolTyrKar17}, Filter Pruning (FP)~\cite{LiKadDurEtAl17} and Channel Pruning (CP)~\cite{He2017Channel}.

\subsection{ConvNet on CIFAR-10}
CIFAR-10 dataset contains~$10$ classes with~$50,000$ images for training and~$10,000$ for testing. We take~$5,000$ images from the training set as the validation set. ConvNet is a tiny version of AlexNet for CIFAR-10 dataset, which is composed of~$3$ convolutional layers and~$1$ fully connected layer. The batch size is~$256$ and initial learning rate~$0.001$.

For simplicity, we set the pruning ratios of the $3$ convolutional layers to the same value. The performance of different methods is shown in Tab.\ref{tab:result_convnet}. Consistently, SPP is much better than the other three methods, and especially when the speedup is small, such as$2\times$, SPP can even improve the performance, for which we argue that the modest pruning can take out the redundancy, which makes the model prone to overfitting, regularize the objective function and thus increase the accuracy. This phenomenon was also found by other pruning methods~\cite{WenWuWan16}. 


\begin{table}[!h]
    \newcommand{\tabincell}[2]{\begin{tabular}{@{}#1@{}}#2\end{tabular}}
    \centering
    \begin{tabular}{lccccc}
       \toprule
         \multirow{2}*{Method}  &  \multicolumn{5}{c}{Increased err. (\%)} \\
         \cline{2-6}
                       & $2\times$ & $4\times$ & $6\times$ & $8\times$ & $10\times$ \\
         \midrule
         TP~\cite{MolTyrKar17} (our impl.)      &  $1.0$ & $3.4$ & $5.4$ & $7.1$ & $8.7$ \\
         FP~\cite{LiKadDurEtAl17} (our impl.)   &  $1.6$ & $3.6$ & $5.0$ & $7.5$ & $8.5$ \\
         SSL~\cite{WenWuWan16}                 &  $3.0$ & $5.8$ & $7.4$ & $7.5$ & $8.3$ \\
         Ours                                  & -$\mathbf{0.2}$ & $\mathbf{0.3}$ & $\mathbf{1.2}$ & $\mathbf{1.5}$ & $\mathbf{2.4}$ \\
       \bottomrule
    \end{tabular}
     \caption{The increased error of different pruning methods when accelerating ConvNet on CIFAR-10. The baseline test accuracy is $81.5\%$. Minus means the test accuracy is improved.}
    \label{tab:result_convnet}
 \end{table}

\textbf{Recovery Analysis.}
To demonstrate the recovery effect of SPP, we studies the fraction of weights which are not important at the beginning, but become important at the final stage of training. We calculate the fraction of weights whose ranks are below~$RN_c$ at the beginning and then above~$RN_c$ at the final stage. These weights should be pruned by many one-shot pruning methods, but they are finally retained by SPP. Tab.\ref{tab:recovery} shows that this fraction, termed as \emph{recovery ratio}, for three convolutional layers of ConvNet. Because the first layer~(\verb+conv1+) is relatively small, consisting of only~$75$ columns, we think it may lack the dynamics which is necessary to recover misjudgments, thus none of them recovered. However, for \verb+conv2+ and \verb+conv3+, which consist of~$800$ columns each, the recovery ratios are very prominent, indicating that SPP could effectively utilize the dynamics of model parameters and achieve good performance. This may explain why the performance of SPP is more robust for large pruning ratios than the other three methods.

\begin{table}
    \newcommand{\tabincell}[2]{\begin{tabular}{@{}#1@{}}#2\end{tabular}}
    \centering
    \begin{tabular}{lccccc}
       \toprule
         Layer  & $2\times$ & $4\times$ & $6\times$ & $8\times$ & $10\times$\\
         \midrule
         conv1 & $0$ & $0$ & $0$ & $0$ & $0$ \\
         conv2 & $7.9\%$ & $14.0\%$ & $16.6\%$ & $25.2\%$ & $32.7\%$ \\
         conv3 & $11.4\%$ & $17.3\%$ & $16.9\%$ & $22.9\%$ & $30.8\%$ \\

       \bottomrule
    \end{tabular}
     \caption{Recovery ratios for the three convolutional layers of ConvNet, with different speedup ratios.}
    \label{tab:recovery}
 \end{table}




\subsection{AlexNet on ImageNet}
\label{section:alexnet_imagenet}
We further verify our method on ImageNet-2012, which is a large dataset of~$1,000$ classes, containing~$1.28$M images for training, $100,000$ for testing and~$50,000$ for validation. There are~$5$ convolutional layers and~$3$ fully connected layers in AlexNet~\cite{KriSutHin12}. Here we adopt the CaffeNet, an open re-implementation of AlexNet, as the pre-trained model. The baseline top-5 accuracy is $80.0\%$ on ImageNet-2012 validation set.

\begin{figure}[h]
   \begin{minipage}{0.5\linewidth}
      \centering
      \includegraphics[width=\textwidth]{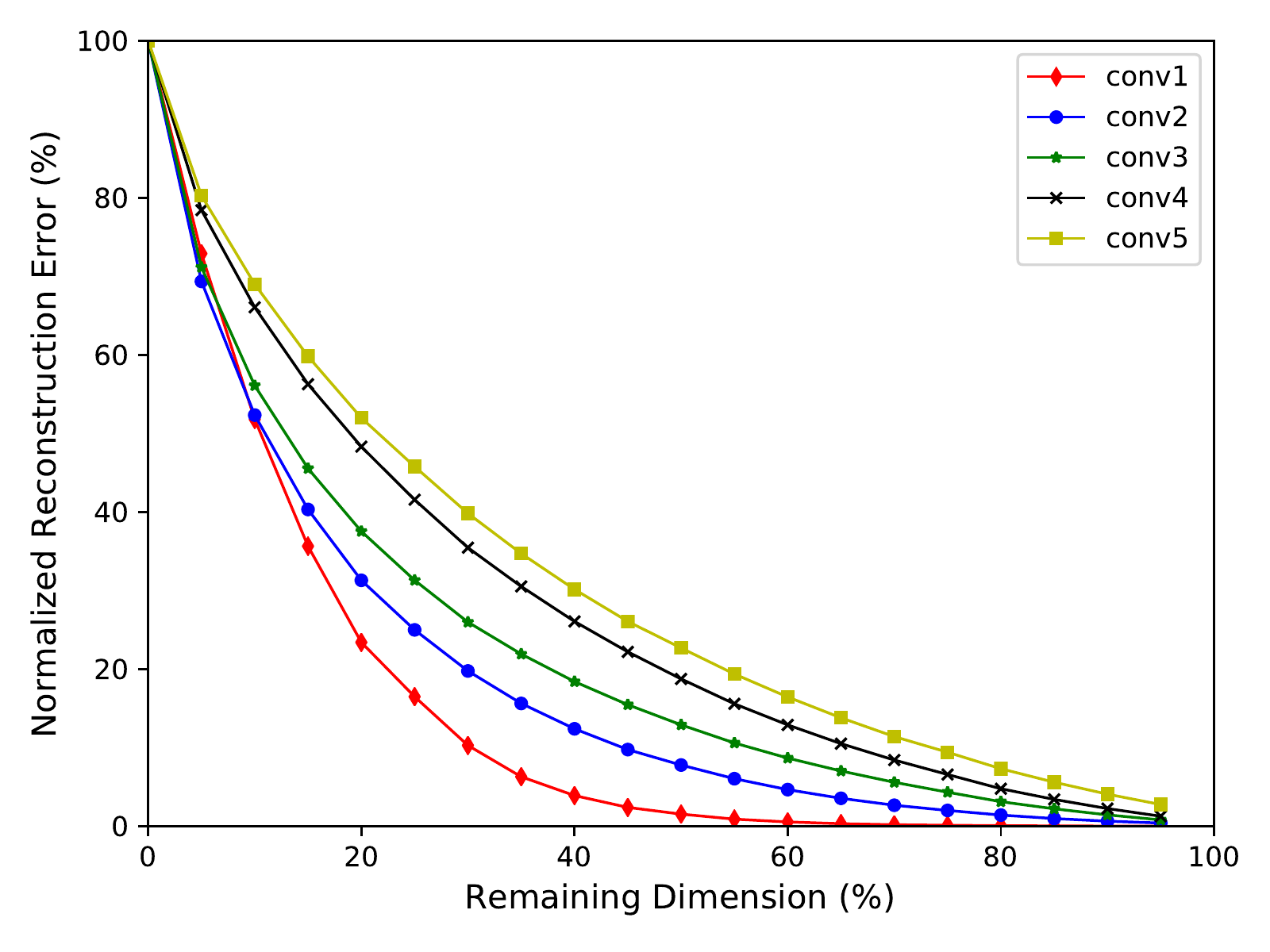}
   \end{minipage}
   \begin{minipage}{0.5\linewidth}
      \centering
      \includegraphics[width=0.9\textwidth, height=0.21\textheight]{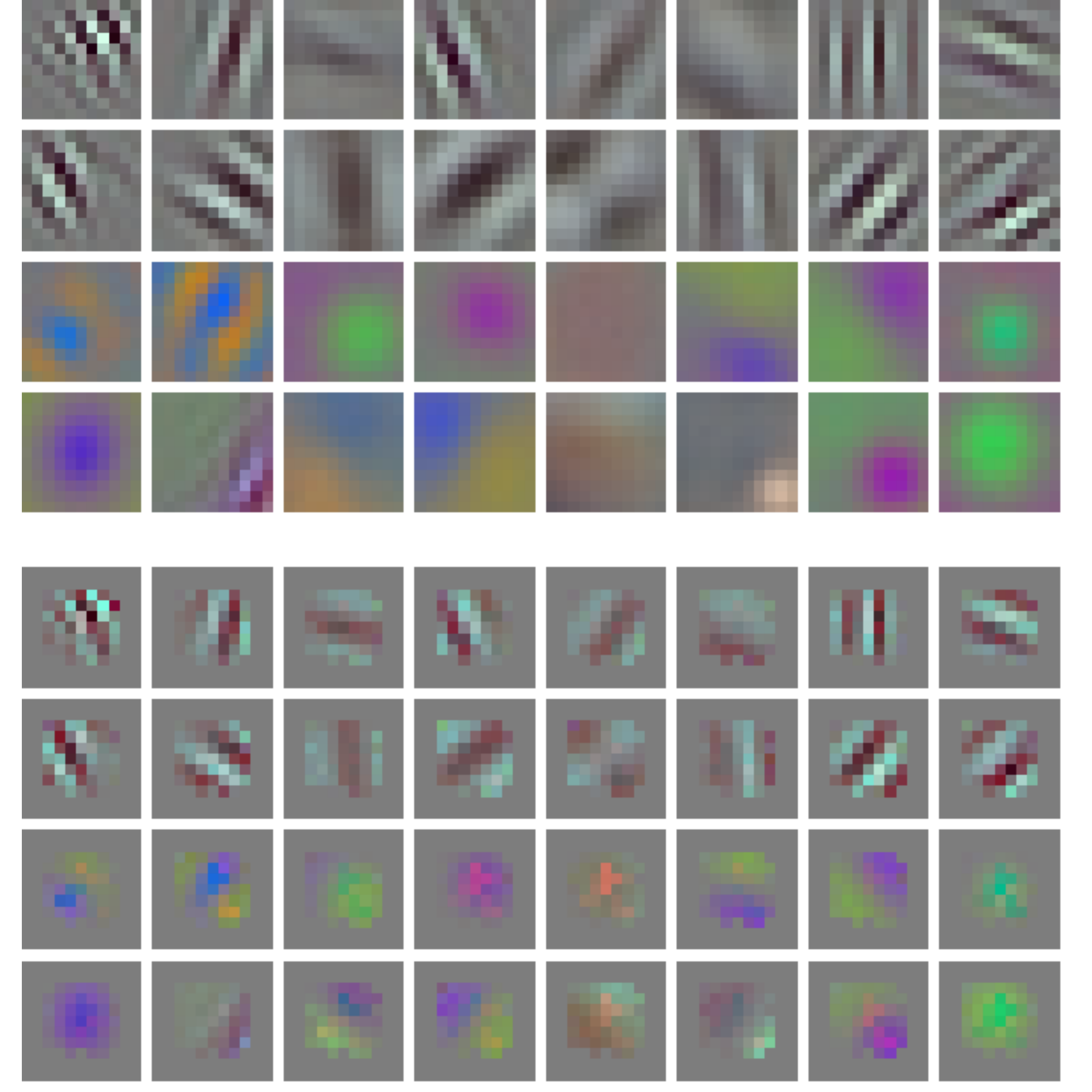} 
   \end{minipage}
   \caption{\textbf{left}: The normalized reconstruction errors with different remaining PCA ratios for the five convolutional layers in AlexNet. \textbf{right}: Filter visualization of first conv layer. The first four rows are the filters from the unpruned AlexNet, the following four rows are corresponding counterparts after SPP pruning with $4\times$ speedup, where pruned weights are illustrated by the gray blocks in each filter (\emph{better seen when in color and zoomed in}).}
   \label{fig:alexnet-sensitivity_conv1-visualization}
\end{figure}

Because AlexNet, VGG-16 and ResNet-50 are much deeper CNN models, where constant pruning ratio for all layers is not optimal. Usually pruning ratios of different layers are determined according to their sensitivity. There are mainly two ways to evaluate sensitivity: (1) Fix other layers and prune one layer, using the accuracy of the pruned model as the sensitivity measure for the pruned layer~\cite{LiKadDurEtAl17}; (2) Apply PCA to each layer and take the reconstruction error as the sensitivity measure~\cite{He2017Channel,WenWuWan16}. Here we take the PCA approach for simplicity. The PCA analysis of AlexNet is shown in Fig.\ref{fig:alexnet-sensitivity_conv1-visualization} (left). We plot the normalized reconstruction errors with different remaining principle component ratios. It can be seen that, under the same remaining principle component ratio, the normalized construction error of upper layers (like \verb+conv5+) are greater than that of the bottom layers (like \verb+conv1+), which means that the upper layers are less redundant. Thus, we set the proportion of the remaining ratios~(one minus the pruning ratio) of these five layers to~$1:1:1:1.5:1.5$.

Accuracy comparison of TP, FP, CP and SPP is shown in~Tab.\ref{tab:result_alexnet}. SPP outperforms the other three methods by a large margin. SPP can accelerate AlexNet by $5\times$ speedup with only $0.9\%$ increase of top-5 error. And similar to the result of ConvNet on CIFAR-10, when the speedup ratio is small ($2\times$), SPP can even improve the performance (by $0.7\%$). Previous work~\cite{WenWuWan16} reported similar improvements, while their improvement is relatively small ($0.1\%$) and under less speedup settings~($1.3\times$).

\begin{table}
    \centering
    \begin{tabular}{lccc}
       \toprule
         Method  & $2\times$ & $4\times$ & $5\times$ \\
         \midrule
         TP~\cite{MolTyrKar17}                     & $3.9$  & $9.2$ & $13.9$  \\
         FP~\cite{LiKadDurEtAl17} (our impl.)      & $0.6$  & $4.1$ & $4.7$   \\
         SSL~\cite{WenWuWan16}                     & $1.3$  & $4.3$ & $5.3$   \\
         Ours                   & -$\mathbf{0.7}$ & $\mathbf{0.3}$ & $\mathbf{0.9}$   \\
       \bottomrule
    \end{tabular}
     \caption{Accelerating AlexNet on ImageNet. The baseline top-5 accuracy of the original network is $80.0\%$. We carefully tune the speedup ratio to be the same, and compare the top-5 error increase~(\%) of SPP, TP, FP and SSL. }
    \label{tab:result_alexnet}
 \end{table}

\textbf{Visualization Analysis.} To take a closer look at the effect of SPP, we visualize the filters of layer \verb+conv1+ of AlexNet. \verb+conv1+ contains 64 filters of size~$11 \times 11 \times 3$. In Fig.\ref{fig:alexnet-sensitivity_conv1-visualization} (right), we randomly visualize 32 filters as 32 RGB images with size~$11 \times 11$.  Notably, SPP prunes the peripheral parts of filters, while the most salient parts are retained after pruning.

\subsection{VGG-16 on ImageNet}
\label{section:vgg16}
VGG-16 is a deep single-branch convolutional neural network with $13$ convolutional layers. We use the open pre-trained model, whose single-view top-5 accuracy is $89.6\%$.

Like the experiment with AlexNet, we firstly use PCA to explore the redundancy of different layers. Similarly, we found that  deeper layers have less redundancy, in line with previous work~\cite{He2017Channel}. For convenience of comparison, we just follow~\cite{He2017Channel} by setting the proportion of remaining ratios of shallow layers (\verb+conv1_x+ to \verb+conv3_x+), middle layers (\verb+conv4_x+) and top layers (\verb+conv5_x+) to~$1:1.5:2$. Given the computation contribution of layer \verb+conv1_1+ and \verb+conv5_3+ to total computation is relatively small ($0.6\%$ and $3.0\%$ respectively), we choose not to prune these two layers. We firstly prune the network with batch size $64$ and learning rate $0.0005$. Then we retrain the pruned model with batch size~$256$ and learning rate~$0.005$.



 Tab.\ref{tab:result_vgg16} shows the accuracy comparison of TP, FP, CP and SPP. Generally, SPP and CP are much better than FP and TP. Under small speedup ratio ($2\times$), both SPP and CP can achieve zero increase of top-5 error. When the speedup ratio is greater ($4\times$ and $5\times$), the performance of SPP is comparably with CP.

\begin{table}
	\begin{minipage}{0.5\linewidth}
	    \centering
	    \begin{tabular}{lccc}
	       \toprule
	         Method  & $2\times$ & $4\times$ & $5\times$  \\
	         \midrule
	         TP~\cite{MolTyrKar17}  & $-$ & $4.8$ & $-$   \\
	         FP~\cite{LiKadDurEtAl17} (\cite{He2017Channel}'s impl.) & $0.8$ & $8.6$ & $14.6$ \\
	         CP~\cite{He2017Channel}  & $\mathbf{0}$ & $1.0$ & $\mathbf{1.7}$ \\
	         SPP & $\mathbf{0}$ & $\mathbf{0.8}$ & $2.0$ \\
	       \bottomrule
	    \end{tabular}
	     \caption{Acceleration of VGG-16 on ImageNet. The baseline top-5 accuracy of the original network is $89.6\%$. The top-5 error increases~(\%) of TP, FP, CP and SPP are compared.}
	    \label{tab:result_vgg16}
	\end{minipage}
	\begin{minipage}{0.5\linewidth}
	    \centering
	    \begin{tabular}{lcc}
	       \toprule
	         Method  & CPU time (ms)  \\
	         \midrule
	         VGG-16 baseline         & $1736.8$ \\
	         SPP ($2\times$)  & $926.6 (1.9\times)$ \\
	         SPP ($4\times$)  & $548.2 (3.2\times)$ \\
	         SPP ($5\times$)  & $406.6 (4.3\times)$ \\
	       \bottomrule
	    \end{tabular}
	     \caption{Actual speedup of VGG-16. The CPU time is obtained by forward processing one image of size $224\times224\times3$. Experiments are conducted on Intel Xeon(R) CPU E5-2620 v4 @ 2.10GHz with single thread.}
	    \label{tab:actual_speedup}
		\end{minipage}
 \end{table}

\textbf{Actual Speedup Analysis.} We further evaluate the actual speedup of pruned VGG-16 model on CPU with \verb+Caffe+. Results are averaged from 50 runs with batch size~$32$. From Tab.\ref{tab:actual_speedup}, we can see that the pruned model by SPP can achieve actual acceleration on general platform with off-the-shelf libraries. The discrepancy between theoretical speedup (measured by GFLOPs reduction) and actual speedup (measured by inference time) is mainly because of the influence of memory access, the unpruned fully-connected layers  and the non-weight layers such as pooling and ReLU. Note that given the specific implementation differences, these values may vary with different platforms~\cite{MolTyrKar17}.

\subsection{ResNet-50 on ImageNet}
\label{section:resnet50}
Unlike single-branch AlexNet and VGG-16, ResNet-50 is a more compact CNN with multi-branches, which is composed of $53$ convolutional layers. We use the open pre-trained caffemodel, whose top-5 accuracy on ImageNet-2012 validation set is $91.2\%$. Considering ResNet-50 has many layers, we do not set different pruning ratios for different layers. Instead we set the pruning ratio of all convolutional layers to the \emph{same} value in this experiment for simplicity. Pruning batch size is~$64$ and learning rate~$0.0005$. When retraining the batch size is~$256$ and initial learning rate~$0.001$.



The result is shown in Tab.\ref{tab:result_resnet50}. It can be seen that our method achieves better result than CP under~$2 \times$ speedup. Note that the implementation of SPP on ResNet-50 is very simple, just the same as previous experiments. However, given the multi-branch structure of ResNet, CP needs to add the multi-branch enhancement procedure to generalize their method to ResNet.

 \begin{table}
	\begin{minipage}{0.5\linewidth}
	    \centering
	    \begin{tabular}{lc}
	       \toprule
	         Method  & Increased err. (\%) \\
	         \midrule
	         CP (enhanced)~\cite{He2017Channel}  & $1.4$ \\
	         SPP & $\mathbf{0.8}$ \\
	       \bottomrule
	    \end{tabular}
	     \caption{Accelerating ResNet-50 on ImageNet. The baseline top-5 accuracy of the original network is $91.2\%$. Top-5 error increases of CP and SPP are compared under $2\times$ speedup.}
	    \label{tab:result_resnet50}
	\end{minipage}
	\begin{minipage}{0.5\linewidth}
	    \centering
	    \begin{tabular}{lcccc}
	       \toprule
	         Method  &  $2\times$ & $4\times$ & $8\times$ & $16\times$ \\
	         \midrule
	         TP~\cite{MolTyrKar17} (our impl.) & $0.9$          & $3.0$          & $4.5$          & $7.3$ \\
	         SPP            & $\mathbf{0}$   & $\mathbf{1.2}$ & $\mathbf{3.9}$ & $\mathbf{5.5}$ \\
	       \bottomrule
	    \end{tabular}
	     \caption{Comparison of SPP and TP on the Flower-102 dataset. The baseline accuracy is $93.2\%$ and the results are test error increases~(\%) under different speedup ratio.}
	    \label{tab:result_alexnet_flower}
	\end{minipage}
 \end{table}

\subsection{Transfer Learning}
Finally, we apply SPP to transfer learning, where a well-trained model is finetuned by the data from other knowledge domains. We use pre-trained AlexNet model to finetune the Oxford Flower-102 dataset~\cite{NilZis08}. The 102-class dataset is composed of~$8,149$ images, among which $6,149$ is used for training, $1,020$ for validation and the other $1,020$ for testing. We use the open pre-trained caffemodel from Caffe Model Zoo as baseline, whose test accuracy is~$93.2\%$. We compare the performance of our method with TP~\cite{MolTyrKar17}, which was reported very effective for transfer learning.

Like the above experiments, we firstly prune the pre-trained model with SPP, then finetune the model to regain accuracy. The learning rate is set to be~$0.001$ and batch size~$50$. For simplicity, we use a constant pruning ratio in this experiment. Tab.\ref{tab:result_alexnet_flower} shows that SPP consistently outperforms TP at all speedup ratios. Note that in \cite{MolTyrKar17}, the authors claim that Taylor-based criteria was significantly better than $L_1$~norm in pruning for the transfer learning task,while our result shows that~\textbf{SPP+$L_1$~norm} outperforms~\textbf{TP+Taylor-based criteria}, which proves the effectiveness of SPP in transfer learning tasks.

\section{Conclusions}
We proposed Structured Probabilistic Pruning (SPP) for CNN acceleration, which prunes weights of CNN in a probabilistic manner and able to correct misjudgment of importance in early training stages. The effectiveness of SPP is proved by comparison with state-of-the-art methods on popular CNN architectures. Further, the effectiveness of SPP is also demonstrated on transfer learning tasks. 

The effectiveness of SPP shows a more dynamic and smoothed pruning process is worth exploring, but for now it is more heuristic than theoretically rigorous. In the future, we plan to introduce the Bayesian inference idea~\cite{molchanov2017variational,neklyudov2017structured,louizos2017bayesian} into SPP to find a more solid foundation. In addition, it's also interesting to generalize our method to more neural network architectures such as Recurrent Neural Networks.

~\\
\noindent \textbf{Acknowledgments}. This work is supported by the Natural Key R\&D Program of China (Grant No. 2017YFB1002400), Natural Science Foundation of Zhejiang
Province (Grant No. LY16F010004) and Zhejiang Public Welfare Research Program (Grant No. 2016C31062).

\bibliography{ref}
\end{document}